\documentclass[letterpaper]{article} 
\usepackage{aaai2027}  
\usepackage[hyphens]{url}  
\usepackage{graphicx} 
\usepackage{natbib}  
\usepackage{caption} 
\usepackage{algorithm}
\usepackage{algorithmic}
\usepackage{newfloat}
\usepackage{listings}
\DeclareCaptionStyle{ruled}{labelfont=normalfont,labelsep=colon,strut=off} 
\floatstyle{ruled}
\newfloat{listing}{tb}{lst}{}
\floatname{listing}{Listing}

\usepackage{booktabs}
\usepackage{amsmath}
\usepackage{amssymb}
\usepackage{tikz}
\usepackage{enumitem}
\usetikzlibrary{arrows.meta,positioning,fit,calc,backgrounds}

\nocopyright

\title{Universal Concept Disruption for SAM3 Image Segmentation}
\author{
Hao Wang\textsuperscript{\rm 1},
Yuxuan Zhang\textsuperscript{\rm 2}\textsuperscript{\ensuremath{\dagger}},
Wei Yang\textsuperscript{\rm 1,\rm 3,\rm 4}\textsuperscript{\ensuremath{\dagger}}
}
\affiliations{
\textsuperscript{\rm 1}School of Computer Science and Technology, University of Science and Technology of China, Hefei, China\\
\textsuperscript{\rm 2}School of Artificial Intelligence and Computer Science, Jiangnan University, Wuxi, China\\
\textsuperscript{\rm 3}Suzhou Institute for Advanced Research, University of Science and Technology of China, Suzhou, China\\
\textsuperscript{\rm 4}Hefei National Laboratory, Hefei, China\\
wanghao0204@mail.ustc.edu.cn, yxzhang97@jiangnan.edu.cn, qubit@ustc.edu.cn
}

\begin{document}

\maketitle
{\renewcommand{\thefootnote}{\ensuremath{\dagger}}\footnotetext{Corresponding authors}}

\begin{abstract}
SAM3 extends promptable segmentation from geometry-driven mask prediction to open-vocabulary concept segmentation, where a text-conditioned grounding model decides whether a concept is present and segments all matching instances. While this presence-gated design improves concept-level prediction, its adversarial robustness remains unexplored. In this paper, we introduce Universal Concept Disruption (UCD), the first universal cross-concept adversarial attack tailored to SAM3 image segmentation. UCD learns a single bounded image perturbation from (image, noun-phrase) pairs and attacks SAM3 as an integrated concept-grounding system. It jointly disrupts the text-conditioned input path, maximizes divergence in prompt-shared visual features, suppresses the final presence-gated concept scores, and corrupts the spatial validity of retained masks through area collapse and clean-mask Dice disruption. Across SACo-Gold, LVIS, RefCOCO, PhraseCut, and OpenImages datasets, UCD consistently outperforms all baselines under a matched evaluation protocol, reducing average mask AP from 59.43 to 18.73 and average cgF1 from 50.32 to 20.49. The learned perturbation also transfers to SAM3.1 and to SAM3 video inference without re-optimization, while prompt ensembling, lightweight head fine-tuning, and temporal filtering provide limited recovery.
\end{abstract}

\section{Introduction}

Driven by their exceptional zero-shot generalization capabilities, the Segment Anything Model (SAM) series has emerged as a crucial bridge between visual foundation models and diverse downstream perception tasks. Specifically, SAM~\cite{kirillov2023segment} provides a robust framework for prompt-driven image segmentation, and SAM2~\cite{ravi2025sam} advances this paradigm by incorporating temporal awareness for seamless video segmentation and object tracking. The recently released SAM3~\cite{carion2025sam} changes the operating point. It introduces open-vocabulary concept segmentation, and the model is expected to determine whether the concept exists and segment all matching instances. 

Adversarial attacks introduce visually imperceptible perturbations to images, causing the model to produce incorrect predictions and fail at its intended task, thereby achieving the attack objective. The vulnerability of the SAM family to such attacks has been experimentally validated~\cite{long2025robust,qiao2023robustness}. Specifically, methods such as DarkSAM~\cite{zhou2024darksam} and AttackSAM~\cite{zhang2023attack} are designed to target the original SAM. For SAM2, UAP-SAM2~\cite{zhou2026vanish} utilizes universal adversarial perturbations; in scenarios where the initial $k$ frames of a video are continuously attacked, it drastically degrades SAM2's video segmentation performance. However, the robustness of SAM3 against adversarial attacks remains unexplored.

We first verify whether existing attacks can be directly transferred to SAM3 image segmentation. To make the comparison fair, we adapt representative SAM, SAM2, and general adversarial baselines into the same universal-perturbation~\cite{2017Universal} setting, using the same train-test separation, random seeds, optimization budget, and evaluation pipeline. Across five datasets, attacks that do not directly target SAM3's concept-scoring mechanism produce small or inconsistent degradation, while even the strongest adapted baseline remains below our SAM3-specific attack. Existing universal attacks against SAM aim to make the model segment nothing or segment incorrectly across prompts. SAM2 attacks further account for video memory and cross-frame prompt reuse. However, SAM3 differs significantly from SAM1 and SAM2, being specifically tailored for concept segmentation. To enable its concept segmentation capabilities, SAM3 features a highly integrated architecture design on both the input and output sides. A failure is no longer determined only by whether a mask is geometrically distorted, it can also occur when the queried concept is absent and suppresses otherwise plausible candidates.

On the input side, the original SAM conducted a proof-of-concept for text prompts by directly extracting embeddings via a frozen CLIP~\cite{radford2021learning} text encoder. However, its performance was highly suboptimal, which led to subsequent decoupled designs that relied on cascading external CLIP-based detectors to provide spatial prompts. In contrast, SAM3 integrates Perception Encoder (PE)-style~\cite{bolya2026perception} vision-language representations into a unified text-conditioned grounding architecture and is trained end-to-end on the SA-Co~\cite{carion2025sam} dataset. On the output side, SAM3 introduces a presence token to address the challenges of open-vocabulary segmentation. This mechanism incorporates a presence token for concept existence and computes a final candidate score that combines query confidence with presence confidence. Functioning as an existence switch, the presence token is globally multiplied by the scores of all individual masks. While this design enhances the robustness of open-vocabulary segmentation, it inadvertently exposes a novel and critical universal attack surface.

Motivated by this observation, we propose Universal Concept Disruption (UCD), the first cross-concept universal adversarial attack tailored to SAM3 image segmentation. UCD treats SAM3 as a concept-grounding system rather than a promptable mask decoder, and disrupts it at three coupled levels. At the input level, UCD learns a single bounded perturbation from (image, noun-phrase (NP)) pairs, thereby activating the same text-conditioned concept path used at inference instead of relying on geometric prompts. At the feature level, UCD perturbs the prompt-shared visual representation, turning the common image backbone into a
transferable attack carrier across concepts, datasets, and model variants. At the output level, UCD targets SAM3's presence-gated concept decision together with the spatial validity of retained masks, causing present concepts to disappear or become geometrically unreliable. We conduct extensive experiments on five datasets, showcasing superior attack efficacy of UCD. Our contributions are threefold:

\begin{itemize}

\item We propose the first universal cross-concept adversarial attack, termed UCD, specifically targeting the open-vocabulary image segmentation capabilities of SAM3.

\item We develop a SAM3-specific concept disruption framework that couples text-conditioned input optimization, prompt-shared feature disruption, and presence-gated output corruption.

\item We comprehensively evaluate our method across five datasets under a unified universal-perturbation protocol. The results demonstrate that our approach consistently surpasses existing attacks.
\end{itemize}

\section{Methodology}

\subsection{Task Definition and Overview}

\subsubsection{SAM3 Promptable Concept Segmentation}

We focus on the single-image, text-prompted setting. Let $x \in [0,1]^{3 \times H \times W}$ be an image and let $t$ be a text prompt. SAM3 encodes the image and text, then runs a grounding stage that returns a set of candidate boxes, masks, query logits, and a prompt-level presence logit. We denote the image backbone by $E_{\theta}^{\mathrm{img}}$, the text encoder by $E_{\theta}^{\mathrm{text}}$, and the grounding head by $G_{\theta}$. The simplified image inference path is
\begin{equation}
  F = E_{\theta}^{\mathrm{img}}(x), \quad
  q = E_{\theta}^{\mathrm{text}}(t), \quad
  (O,p) = G_{\theta}(F, q),
\end{equation}
where $O=\{(\ell_i, b_i, m_i)\}_{i=1}^{N}$ and $p$ is the prompt-level presence logit. Here $\ell_i$ is the raw query logit for candidate $i$, $b_i$ is its box, and $m_i$ is its mask logit map. For each candidate, the final concept score is the product of the query probability and the presence probability:
\begin{equation}
  s_i(x,t) = \sigma(\ell_i(x,t)) \cdot \sigma(p(x,t)).
  \label{eq:combined-score}
\end{equation}
Post-processing keeps candidates whose $s_i$ exceeds a threshold $\tau$ and converts retained mask logits into binary masks. A candidate with high mask quality can still be discarded if the model predicts the concept as absent, because the presence probability globally gates all candidate scores.

\subsubsection{Universal Concept Attack against Images}

Targeting images, we attack SAM3 with a single additive perturbation $\delta$ shared across all images and prompts:
\begin{equation}
  x^{\mathrm{adv}} = \Pi_{[0,1]}\left(x+\delta\right), \quad
  \|\delta\|_{\infty} \leq \epsilon,
  \label{eq:adv-input}
\end{equation}
where $\epsilon$ denotes the bound on the perturbation magnitude. The attacker has white-box access to the released SAM3 model and optimizes $\delta$ over public training samples. 

We consider an untargeted objective. In SAM3, concept segmentation prompts are categorized into positive prompts (where the concept is present in the image) and negative prompts (where the concept is absent from the image). For positive prompts, the attack aims to remove or corrupt the correctly predicted concept masks. For negative prompts, however, applying the same loss function as that used for positive prompts is ill-posed.

For a training set $\mathcal{D}=\{(x_j,t_j)\}_{j=1}^{M}$ of positive prompts, the ideal universal attack objective can be formulated as minimizing the retained clean-prediction utility:
\begin{equation}
  \min_{\|\delta\|_{\infty}\leq\epsilon}
  \mathbb{E}_{(x,t)\sim \mathcal{D}}
  \left[
  \mathcal{U}\left(f_{\theta}(x^{\mathrm{adv}},t), f_{\theta}(x,t)\right)
  \right],
  \label{eq:generic-objective}
\end{equation}
where $\mathcal{U}$ measures the remaining agreement or utility of the adversarial prediction relative to the clean prediction. Minimizing $\mathcal{U}$ induces adversarial failure under the clipped input $x^{\mathrm{adv}}$ from Eq.~(\ref{eq:adv-input}).

\begin{figure*}[t]
\centering

\includegraphics[width=0.92\textwidth]{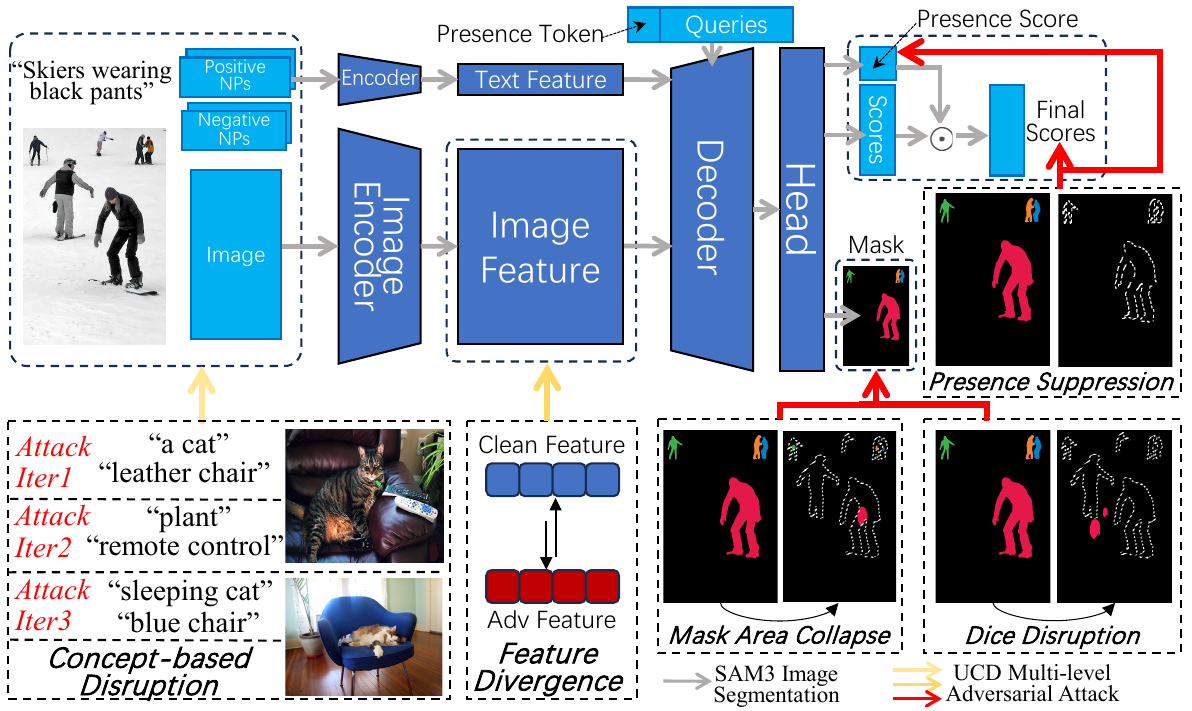}

\caption{Universal Concept Disruption attacks SAM3 at input, feature, and output levels while optimizing a single image-space perturbation shared across prompts.}
\label{fig:ucd-architecture}
\end{figure*}

\subsection{Universal Concept Disruption}

In SAM3, a concept query passes through a hierarchy of decisions:
\begin{itemize}[leftmargin=1.35em,itemsep=1pt,topsep=3pt,parsep=0pt,label=\raisebox{0.15ex}{\scriptsize$\blacktriangleright$}]
    \item The image must produce features that match the NPs.
    \item The grounding head must assign high candidate scores under the global presence gate.
    \item Retained candidates must have useful mask geometry.
\end{itemize}
A perturbation may disturb low-level image features without changing the final concept decision. It may suppress a score while leaving a surviving mask spatially correct, or it may alter a mask while the presence-gated score remains high enough for post-processing.

As shown in Figure~\ref{fig:ucd-architecture}, the attack has three architectural levels. First, at the input-concept level, UCD trains on actual positive image-NP pairs rather than geometric prompt grids. This activates SAM3's text-conditioned grounding path during optimization and avoids reducing the task to prompt-agnostic mask failure. Second, at the feature level, UCD maximizes the discrepancy between clean and adversarial image features. Because the visual backbone is shared across text prompts, this term encourages transfer beyond the specific concepts observed during training. Third, at the output level, UCD directly attacks the presence-gated concept decision and the geometry of the retained masks. The score term targets the final score in Eq.~(\ref{eq:combined-score}), while the mask-area and clean-mask Dice terms make high-score survivors spatially useless. This hierarchy focuses UCD on the disappearance and spatial corruption of present concepts. Activating absent concepts is a complementary false-positive threat model, but it is not the objective studied in this paper. The method below therefore keeps the attack aligned with positive-prompt concept segmentation: suppress the final concept decision, collapse or displace the retained masks, and disturb prompt-shared visual features.

For every sampled image, we first compute clean image features and clean top masks without gradient. We then apply $\delta$, run the same text prompts through SAM3, and back-propagate through the image backbone and grounding head into $\delta$. Let $\mathcal{X}_{+}$ be the set of training images and let $\mathcal{T}_{+}(x)$ be the positive noun phrases associated with image $x$. The full UCD objective is
\begin{equation}
\begin{aligned}
  \mathcal{L}_{\mathrm{out}}
  =&\;
  \lambda_s\mathcal{L}_{\mathrm{score}}+\lambda_a\mathcal{L}_{\mathrm{area}}+\lambda_d\mathcal{L}_{\mathrm{dice}},\\
  \bar{\mathcal{L}}_{\mathrm{out}}(x,\delta)
  =&\;
  \frac{1}{|\mathcal{T}_{+}(x)|}
  \sum_{t\in\mathcal{T}_{+}(x)}
  \mathcal{L}_{\mathrm{out}}(x,t,\delta),\\
  \mathcal{J}(\delta)
  =&\;
  \mathbb{E}_{x\sim\mathcal{X}_{+}}
  \big[
  \bar{\mathcal{L}}_{\mathrm{out}}(x,\delta)
  -\lambda_f\mathcal{D}_{\mathrm{feat}}(x,\delta)
  \big],\\
  \delta^\star
  =&\;
  \arg\min_{\|\delta\|_{\infty}\leq\epsilon}
  \mathcal{J}(\delta).
\end{aligned}
\label{eq:full-loss}
\end{equation}

\subsubsection{Presence Suppression}

Let $S(x,t)=\{s_i(x,t)\}_{i=1}^{N}$ be the final SAM3 concept scores from Eq.~(\ref{eq:combined-score}). With $\mathcal{I}_{s}=\mathrm{TopK}(S(x^{\mathrm{adv}},t),k)$, we penalize high scores through
\begin{equation}
  \mathcal{L}_{\mathrm{score}}(x,t,\delta)
  =
  \frac{1}{k}
  \sum_{i\in\mathcal{I}_{s}}
  [s_i(x^{\mathrm{adv}},t)-m_s]_{+}.
  \label{eq:score-loss}
\end{equation}
Here $[z]_{+}=\max(z,0)$ and $m_s$ is a score margin. This term attacks the post-processing threshold rather than an intermediate query logit alone. As a result, it simultaneously accounts for the model's concept existence estimate and candidate confidence.

\subsubsection{Mask Area Collapse}

Score suppression alone can lower confidence while leaving plausible spatial masks. To attack segmentation quality, we add a mask-area loss on the top-scoring candidates. Let $M_i(x,t)=\sigma(m_i(x,t))$ be the soft mask probability for candidate $i$. With $\mathcal{I}_{a}=\mathrm{TopK}(S(x^{\mathrm{adv}},t),k_a)$, where the selected scores are treated as fixed for gradient computation, we minimize
\begin{equation}
  \mathcal{L}_{\mathrm{area}}(x,t,\delta)
  =
  \frac{1}{k_a}
  \sum_{i\in\mathcal{I}_{a}}
  \frac{1}{HW}\sum_{u,v} M_i^{uv}(x^{\mathrm{adv}},t).
  \label{eq:area-loss}
\end{equation}
This term encourages high-confidence masks to shrink, which is useful when a candidate remains above threshold but no longer covers the target concept.

\subsubsection{Dice Disruption}

The area loss can over-favor tiny masks but does not explicitly push masks away from the clean spatial support of all high-score instances. Therefore, we store a clean high-score mask set
\begin{equation}
  \mathcal{I}_{c}=\mathrm{TopK}(S(x,t),k_d),\quad
  \mathcal{M}_{c}=\{M_j(x,t):j\in\mathcal{I}_{c}\},
\end{equation}
and minimize the best Dice overlap between each top adversarial mask and this clean set. Let $\mathcal{I}_{d}=\mathrm{TopK}(S(x^{\mathrm{adv}},t),k_d)$ and
\begin{equation}
  \mathrm{Dice}(A,B)
  =
  \frac{2\langle A,B\rangle+\eta}
  {\|A\|_{1}+\|B\|_{1}+\eta}.
  \label{eq:soft-dice}
\end{equation}
The Dice disruption loss is
\begin{equation}
\begin{aligned}
  \mathcal{L}_{\mathrm{dice}}(x,t,\delta)
  =
  \frac{1}{k_d}
  \sum_{i\in\mathcal{I}_{d}}
  \max_{j\in\mathcal{I}_{c}}
  \mathrm{Dice}\big(
  M_i(x^{\mathrm{adv}},t),
  M_j(x,t)
  \big).
\end{aligned}
  \label{eq:dice-loss}
\end{equation}
Although Eq.~(\ref{eq:dice-loss}) is written as a loss to minimize, each term is the strongest Dice overlap with any clean high-score instance. Minimizing it lowers spatial agreement with the clean multi-instance prediction. The clean masks are detached so gradients update only the universal perturbation.

\subsubsection{Feature Divergence}

Text prompts vary widely across concept segmentation benchmarks. To reduce overfitting to individual prompts, we also attack the image representation shared across prompts. Let $F_{\ell}(x)$ denote the last feature map in SAM3's image-backbone FPN. We maximize normalized feature divergence:
\begin{equation}
  \mathcal{D}_{\mathrm{feat}}(x,\delta)
  =
  \frac{
  \|F_{\ell}(x^{\mathrm{adv}})-F_{\ell}(x)\|_2^2
  }{
  \|F_{\ell}(x)\|_2^2+\eta
  }.
  \label{eq:feature-loss}
\end{equation}
The negative sign in Eq.~(\ref{eq:full-loss}) maximizes this divergence while all other terms are minimized. Prompt-invariant image features are a natural attack surface for cross-prompt universal perturbations, but it is paired here with SAM3-specific score and mask objectives.

\subsubsection{Optimization}

We optimize $\delta$ with Adam~\cite{kingma2014adam} and project it into the $\ell_{\infty}$ ball after every step:
\begin{equation}
\begin{aligned}
  \tilde{\delta}
  &\leftarrow
  \mathrm{Adam}(\delta,\nabla_{\delta}\mathcal{J}),\ 
  \delta
  \leftarrow
  \mathrm{clip}(\tilde{\delta},-\epsilon,\epsilon).
\end{aligned}
\end{equation}


\section{Experiments}
\subsection{Experimental Setup}

\subsubsection{Datasets and Metrics}

We evaluate UCD on five concept-segmentation datasets. SACo-Gold is released with SAM3 and evaluates promptable concept segmentation with both positive and negative NPs. LVIS~\cite{gupta2019lvis} evaluates open-vocabulary category segmentation with positive and negative category prompts. RefCOCO~\cite{kazemzadeh2014referitgame}, PhraseCut~\cite{wu2020phrasecut}, and OpenImages~\cite{kuznetsova2020open} provide complementary referring-expression, phrase-grounding, and open-vocabulary segmentation evaluations.

We report cgF$_1$ and COCO-style AP for masks and boxes. cgF$_1$ is the official SAM3 concept-grounding metric, but it is not applicable when a sampled set contains only positive prompts. For each image benchmark, we sample 100 images per dataset for evaluation. For randomly sampled evaluation subsets, we repeat the sampling and evaluation procedure three times and report the averaged metrics. Unless otherwise stated, seed 0 is used as the base sampling seed. We use the official pretrained SAM3 model as the target.

\subsubsection{Implementation Details}

The main perturbation is trained on 50 randomly sampled LVIS-train images and 50 randomly sampled SACo-Silver images. For each image, at most four positive NPs are used to form the positive-prompt training set $\mathcal{D}_{+}$. Training is grouped by image: we perform two updates for each sampled image and run one epoch over the sampled training set. The perturbation is randomly initialized and constrained by $\epsilon=8/255$ in normalized pixel units. We use a learning rate of $0.03$. All images are resized to $1008\times1008$ for training and evaluation. During evaluation, candidate masks are kept with a final-score threshold of $0.5$ and binarized with a mask threshold of $0.5$. The loss weights are $\lambda_s= 0.5$, 
  $\lambda_a = 0.15$, 
  $\lambda_d = 0.5$, 
$\lambda_f= 1$. For the TopK, we use $k=10$, $k_a=k_d=3$.
All experiments are conducted on a single NVIDIA RTX A6000 GPU.

\subsection{Comparison Study}

\begin{table*}[t]
\centering
\small
\setlength{\tabcolsep}{3.5pt}

\resizebox{\textwidth}{!}{%
\begin{tabular}{lcccccccccc}
\toprule
& \multicolumn{6}{c}{Mask AP $\downarrow$} & \multicolumn{1}{c}{Box AP $\downarrow$} & \multicolumn{3}{c}{cgF$_1$ $\downarrow$} \\
\cmidrule(lr){2-7}\cmidrule(lr){8-8}\cmidrule(lr){9-11}
Method & SACo-Gold & LVIS & RefCOCO & PhraseCut & OpenImages & Avg. & Avg. & SACo-Gold & LVIS & Avg. \\
\midrule
Clean & 65.68 & 64.00 & 57.30 & 34.34 & 75.85 & 59.43 & 60.35 & 56.43 & 44.21 & 50.32 \\
\midrule
DarkSAM~\cite{zhou2024darksam} & 65.41 & 62.43 & 57.45 & 35.72 & 76.59 & 59.52 & 60.51 & 57.81 & 42.67 & 50.24 \\
S-RA~\cite{shen2024practical} & 61.54 & 61.85 & 57.71 & 34.97 & 76.22 & 58.46 & 59.53 & 51.69 & 42.95 & 47.32 \\
UAPGD~\cite{2020Universal} & 57.00 & 61.43 & 58.09 & 35.12 & 77.09 & 57.75 & 58.56 & 49.57 & 42.42 & 46.00 \\
UAP-SAM2~\cite{zhou2026vanish} & 57.30 & 61.07 & 56.63 & 34.58 & 74.50 & 56.82 & 58.63 & 49.23 & 43.18 & 46.20 \\
CPA~\cite{long2025robust} & 38.34 & 49.76 & 51.91 & 29.54 & 68.15 & 47.54 & 47.40 & 27.94 & 28.93 & 28.44 \\
GRAT~\cite{2024Transferable} & 23.97 & 31.83 & 19.13 & 10.01 & 23.87 & 21.76 & 21.83 & 20.60 & 23.88 & 22.24 \\
\midrule
Ours & \textbf{18.69} & \textbf{30.51} & \textbf{13.77} & \textbf{9.44} & \textbf{21.23} & \textbf{18.73} & \textbf{17.77} & \textbf{17.40} & \textbf{23.57} & \textbf{20.49} \\
\bottomrule
\end{tabular}
}

\caption{Comparison with SAM3-adapted baseline attacks under the same fair universal-perturbation training budget. We report COCO-style mask AP (\%), COCO-style bounding-box AP averaged over five datasets (\%), and concept-aware F1 (cgF$_1$, \%) on datasets with negative prompts. Clean denotes the unattacked SAM3 performance. Lower adversarial values indicate stronger attacks; the best attack values are shown in bold.}
\label{tab:comparison_study}
\end{table*}

Table~\ref{tab:comparison_study} compares UCD with SAM3-adapted baseline attacks. Each baseline is optimized as a universal image perturbation under the same perturbation budget, training data, random seeds, and optimization schedule as UCD, and is evaluated on disjoint test samples.

UCD achieves the strongest attack among all baselines. Averaged over the five benchmarks, UCD reduces mask AP from 59.43 to 18.73, while the strongest baseline, GRAT, remains at 21.76 and CPA remains at 47.54. The same trend holds for localization: UCD lowers the average box AP from 60.35 to 17.77, compared with 21.83 for GRAT. On concept-aware recognition, UCD reduces the average cgF$_1$ from 50.32 to 20.49, with the lowest adversarial cgF$_1$ on both SACo-Gold and LVIS. SAM/SAM2-style and generic universal perturbation baselines, including DarkSAM, S-RA, UAPGD, and UAP-SAM2, produce only small degradation and sometimes fail to reduce AP. Stronger baselines such as CPA and GRAT perform more effectively, but they still fall short of UCD, especially on the cross-dataset mask AP average and box AP average. The per-dataset results further suggest that UCD is not overfitted to the training distribution. The perturbation is trained on a small set of LVIS and SACo-Silver images, yet it achieves better performance on all datasets. This includes datasets with category prompts, referring expressions, phrase-level prompts, and open-vocabulary object annotations. The consistent degradation indicates that jointly attacking feature representations, final concept scores, and mask geometry is more effective.

\subsection{Ablation}

All ablations follow the main evaluation protocol. Mask AP Avg is averaged over
the five image benchmarks, and cgF$_1$ Avg is averaged over SACo-Gold and LVIS.
Unless otherwise specified, Full UCD uses $\epsilon=8/255$.

\subsubsection{UCD Loss Components}

Table~\ref{tab:ablation_loss} shows that all terms contribute to UCD. Removing
feature divergence causes the largest drop in attack strength, while removing
score or mask-geometry losses also weakens the result. Thus, feature disruption
drives transfer, and the output losses align it with SAM3's final concept masks.

\begin{table}[t]
\centering
\small
\setlength{\tabcolsep}{6pt}

\begin{tabular}{lcc}
\toprule
Variant & Mask AP Avg & cgF$_1$ Avg \\
\midrule

w/o score & 36.85 & 33.77 \\
w/o mask area & 46.28 & 38.34 \\
w/o mask dice & 31.01 & 29.48 \\
w/o feature & 57.45 & 44.33 \\
Full UCD & \textbf{18.73} & \textbf{20.49} \\
\bottomrule
\end{tabular}
\caption{Ablation of UCD loss components. Scores are adversarial percentages.}
\label{tab:ablation_loss}
\end{table}

\subsubsection{Concept Target}

Table~\ref{tab:ablation_target} shows that targeting either query confidence or
presence alone is insufficient. Attacking their presence-gated combination gives
the strongest concept-decision disruption.

\begin{table}[t]
\centering
\small
\setlength{\tabcolsep}{6pt}

\begin{tabular}{lcc}
\toprule
Variant & Mask AP Avg & cgF$_1$ Avg \\
\midrule

query-score only & 37.45 & 32.33 \\
presence only & 57.34 & 45.03 \\
query-score + presence & \textbf{18.73} & \textbf{20.49} \\
\bottomrule
\end{tabular}
\caption{Ablation of concept-decision targets. Scores are adversarial
percentages.}
\label{tab:ablation_target}
\end{table}

\subsubsection{Training Prompt Form}

Table~\ref{tab:ablation_prompt} compares text-prompt training with point-grid
training. The sparse and dense grids use $4\times4$ and $8\times8$ points, respectively. Grid supervision still transfers, but text prompts are
clearly stronger, supporting optimization on SAM3's text-conditioned concept
path.

\begin{table}[t]
\centering
\small
\setlength{\tabcolsep}{6pt}

\begin{tabular}{lcc}
\toprule
Variant & Mask AP Avg & cgF$_1$ Avg \\
\midrule

sparse point grid & 27.94 & 29.93 \\
dense point grid & 26.42 & 29.39 \\
text-prompt & \textbf{18.73} & \textbf{20.49} \\
\bottomrule
\end{tabular}
\caption{Ablation of the training prompt form. Evaluation always uses text
prompts. Scores are adversarial percentages.}
\label{tab:ablation_prompt}
\end{table}

\subsubsection{Image-Prompt Pairing}

Table~\ref{tab:ablation_pairing} shows that using more aligned prompts per image
improves the attack. In the shuffled variant, prompt strings are globally
permuted across the sampled positive training items after dataset sampling.
Shuffling prompts remains effective but weaker than the aligned four-prompt
setting, suggesting that both prompt diversity and image-text alignment help.

\begin{table}[t]
\centering
\small
\setlength{\tabcolsep}{6pt}

\begin{tabular}{lcc}
\toprule
Variant & Mask AP Avg & cgF$_1$ Avg \\
\midrule

1 prompt & 46.00 & 39.43 \\
2 prompts & 37.15 & 31.16 \\
3 prompts & 24.91 & 29.72 \\
4 prompts (shuffled) & 27.71 & 30.21 \\
4 prompts & \textbf{18.73} & \textbf{20.49} \\
\bottomrule
\end{tabular}
\caption{Ablation of image-prompt pairing. Scores are adversarial percentages.}
\label{tab:ablation_pairing}
\end{table}

\subsubsection{Perturbation Budget}

Table~\ref{tab:ablation_budget} shows the expected budget-robustness trade-off:
$4/255$ is weak, $8/255$ is the main setting, and $16/255$ further strengthens
the attack.

\begin{table}[t]
\centering
\small
\setlength{\tabcolsep}{6pt}

\begin{tabular}{lcc}
\toprule
Budget & Mask AP Avg & cgF$_1$ Avg \\
\midrule
$4/255$ & 56.73 & 44.05 \\
$8/255$ & 18.73 & 20.49 \\
$16/255$ & \textbf{5.81} & \textbf{7.37} \\
\bottomrule
\end{tabular}
\caption{Ablation of the perturbation budget. The $8/255$ row uses the selected
Full UCD perturbation. Scores are adversarial percentages.}
\label{tab:ablation_budget}
\end{table}

\subsection{Attack Transferability}

\subsubsection{Black-Box Transfer}

We evaluate black-box transfer by training the perturbation on the released
SAM3 image model and applying it directly to
SAM3.1~\cite{carion2025sam} without re-optimization, tuning, or access to
SAM3.1 gradients. Because SAM3.1 keeps the same text-conditioned
concept-segmentation interface and presence-gated prediction semantics, this
setting tests whether UCD transfers beyond one checkpoint. The SAM3.1 evaluation uses the same settings as the main experiments. Table~\ref{tab:sam31_transfer} reports mask AP and cgF$_1$, values
in parentheses are clean-to-adversarial drops. cgF$_1$ is reported for
SACo-Gold and LVIS.

\begin{table}[t]
\centering
\small
\setlength{\tabcolsep}{4pt}

\begin{tabular}{lcccc}
\toprule
\multicolumn{1}{c}{Dataset} & \multicolumn{2}{c}{Mask AP} & \multicolumn{2}{c}{cgF$_1$} \\
\cmidrule(lr){2-3}\cmidrule(lr){4-5}
& Clean & Adv. & Clean & Adv. \\
\midrule
SACo-Gold & 59.87 & 14.43 (-45.44) & 58.97 & 17.51 (-41.46) \\
LVIS & 51.05 & 22.17 (-28.88) & 44.47 & 22.97 (-21.50) \\
RefCOCO & 32.94 & 11.24 (-21.70) & - & - \\
PhraseCut & 26.07 & 7.37 (-18.70) & - & - \\
OpenImages & 58.81 & 13.10 (-45.71) & - & - \\
\midrule
Average & 45.75 & 13.66 (-32.09) & 51.72 & 20.24 (-31.48) \\
\bottomrule
\end{tabular}
\caption{Black-box transfer from SAM3 to SAM3.1. The perturbation is trained on SAM3 and evaluated on SAM3.1 without re-optimization. Parentheses in the adversarial columns denote the drop from clean performance.}
\label{tab:sam31_transfer}
\end{table}

The SAM3-trained perturbation transfers strongly to SAM3.1. Average mask AP
drops from 45.75 to 13.66, and average cgF$_1$ over SACo-Gold and LVIS drops
from 51.72 to 20.24. Since these drops occur without SAM3.1-specific
optimization, they support the claim that prompt-shared image features,
presence-gated scores, and mask geometry form a stable attack surface across
the SAM3 family.

\subsubsection{Image-to-Video Transfer Attack}

We next test cross-task transfer to SAM3 video inference. The perturbation is
trained only with the image objective in Eq.~(\ref{eq:full-loss}) and then
injected into the video frame stream; no video frames, temporal losses, or video
gradients are used during attack training.

Let $v=(x_1,\ldots,x_L)$ denote a video clip and let $t$ be a noun phrase
prompt. SAM3 receives the prompt on the first frame and propagates the resulting
concept tracks. For prefix length $k$, we perturb only the first $k$ frames:
\begin{equation}
  \tilde{x}^{(k)}_i =
  \begin{cases}
  \Pi_{[0,1]}(x_i+\delta), & 1 \leq i \leq k,\\
  x_i, & k < i \leq L.
  \end{cases}
\end{equation}
All later frames are clean. To isolate temporal transfer, we evaluate only the
tail
\begin{equation}
  \mathcal{E}_k=\{k+1,\ldots,L\}.
\end{equation}
For any video metric $M$ where larger is better, we measure the
clean-to-adversarial drop on this tail:
\begin{equation}
  \Delta_M(k)=
  M\!\left(\Phi_\theta(v,t)|_{\mathcal{E}_k}, g|_{\mathcal{E}_k}\right)
  -
  M\!\left(\Phi_\theta(\tilde{v}^{(k)},t)|_{\mathcal{E}_k},
  g|_{\mathcal{E}_k}\right),
\end{equation}
where $\Phi_\theta$ is the SAM3 video predictor and $g$ is the ground-truth
masklet annotation.

\begin{figure}[t]
\centering
\includegraphics[width=\columnwidth]{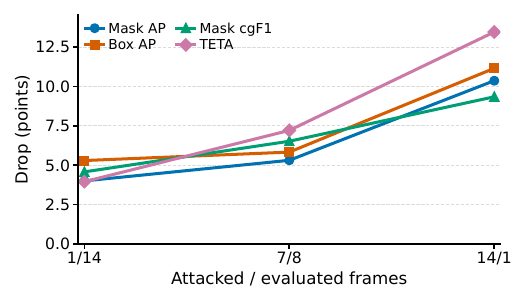}
\caption{Image-to-video transfer on SACo-VEval. A universal perturbation
trained only on images is applied to the first $k$ frames of a 15-frame video
clip, while representative video metrics are computed only on the remaining
clean tail frames. Larger drops indicate stronger temporal transfer.}
\label{fig:image-video-transfer}
\end{figure}

We evaluate on the validation split of SACo-VEval using 15-frame clips and
$k\in\{1,7,14\}$, corresponding to attack/evaluation lengths of $1/14$,
$7/8$, and $14/1$. After filtering short clips, the evaluation contains
50 videos and 200 video-prompt pairs, with at most four positive prompts per video. Figure~\ref{fig:image-video-transfer} plots official SACo-VEval drops on the tail frames.

The image-trained perturbation transfers to video without video-specific
optimization. Attacking only the first frame already reduces tail performance
by 4.00 mask AP, 5.29 box AP, 4.58 mask cgF$_1$, and 3.95 TETA points. Longer
attacked prefixes generally increase the effect: in the $14/1$ setting, the
evaluated frame is clean, yet the drops reach 10.36 mask AP, 11.15 box AP, 9.34
mask cgF$_1$, and 13.46 TETA points. This shows that prefix corruption can be
carried forward by SAM3 video memory to later clean frames.

\subsection{Defense-Oriented Evaluation}
We evaluate three task-relevant mitigation families: prompt ensembling,
adversarial head fine-tuning, and temporal consistency filtering. For each
defense, we report clean performance, fixed-attack performance, and, when the
defense is differentiable or can be approximated, adaptive-attack performance. We avoid using image corruptions as defenses, because they can
reduce clean performance enough to obscure whether the defense is meaningful.

\paragraph{Prompt ensembling.}
SAM3's concept path is text-conditioned, so a natural test-time mitigation is to
query multiple paraphrases of the same noun phrase and aggregate the predictions.
For a prompt $t$, we construct a small deterministic template set
\begin{equation}
  \mathcal{E}(t)=\{t,\ \text{``a photo of }t\text{''},\
  \text{``all }t\text{''},\ \text{``the visible }t\text{''}\}.
\end{equation}
We merge predictions with mask non-maximum suppression and retain candidates whose mean final concept score across matched ensemble members exceeds the standard threshold. This defense tests whether UCD overfits to one lexical realization of the concept and whether voting over prompt variants can recover missed detections.

\paragraph{Adversarial head fine-tuning.}
We evaluate a defense that freezes the
image backbone and text encoder, and fine-tunes only the grounding head and
presence-score layers on a mixture of clean and UCD-perturbed training samples.
For every positive image-prompt pair, the model is trained to recover the clean
concept score and mask assignment under both $x$ and
$\Pi_{[0,1]}(x+\delta)$. Negative prompts are kept in the batch to discourage a
trivial increase of all presence scores. This defense directly targets the
component attacked by UCD while limiting clean-performance drift. To avoid
overstating robustness, we also retrain an adaptive UCD perturbation against the
fine-tuned model and evaluate whether the recovered robustness persists under a
white-box attacker.

\paragraph{Temporal consistency filtering.}
For video inference, we test whether the image-trained perturbation can be
mitigated by rejecting temporally unstable concept tracks. Given frame-level
track scores $s_i$ and masks $M_i$, the filter maintains a track only when its
median score over a sliding window of length $w$ remains above the standard
threshold and its mask remains spatially consistent with the previous accepted
mask:
\begin{equation}
  \begin{aligned}
  \mathrm{keep}(i)=
  \mathbb{1}\big[
  &\mathrm{median}(s_{i-w+1:i})\geq\tau \\
  &{}\wedge\ \mathrm{IoU}(M_i,M_{i-1})\geq \tau_{\mathrm{temp}}
  \big].
  \end{aligned}
\end{equation}
We report metrics on the clean tail frames, as in Fig.~\ref{fig:image-video-transfer}.
This isolates whether temporal post-processing can prevent early corrupted
concept states from propagating into later frames.

\paragraph{Image defenses.}
For image evaluation, we report the same summary metrics used throughout
the ablation study. Table~\ref{tab:defense_image} shows that prompt ensembling is not a
satisfactory defense. Although it lowers the fixed-UCD drop, it also reduces
clean Mask AP Avg from 59.43 to 16.55.
The apparent robustness therefore comes mainly from suppressing predictions on
clean images. Lightweight head fine-tuning preserves clean utility, with 60.62
Mask AP Avg and 48.90 cgF$_1$ Avg, but fixed UCD still reduces performance to 20.84
Mask AP Avg and 17.21 cgF$_1$ Avg. Retraining UCD gives
32.28 Mask AP Avg and 22.71 cgF$_1$ Avg, which is higher than the fixed-attack
score but still below clean performance.

\begin{table}[t]
\centering
\small
\setlength{\tabcolsep}{4pt}

\begin{tabular}{llcc}
\toprule
Defense & Setting & Mask AP Avg & cgF$_1$ Avg \\
\midrule
None & Clean & 59.43 & 50.32 \\
None & Fixed UCD & 18.73 & 20.49 \\
\midrule
Prompt ensemble & Clean & 16.55 & 12.81 \\
Prompt ensemble & Fixed UCD & 2.03 & 4.53 \\
\midrule
Head fine-tuning & Clean & 60.62 & 48.90 \\
Head fine-tuning & Fixed UCD & 20.84 & 17.21 \\
Head fine-tuning & Adaptive UCD & 32.28 & 22.71 \\
\bottomrule
\end{tabular}
\caption{Defense-oriented image evaluation. Scores are percentages. Fixed UCD uses the perturbation from Table~\ref{tab:comparison_study}; Adaptive UCD is retrained against the fine-tuned head.}
\label{tab:defense_image}
\end{table}

\paragraph{Video defense.}
Table~\ref{tab:defense_video} evaluates temporal filtering in the same
image-to-video transfer setting as Fig.~\ref{fig:image-video-transfer}, using
only tail-frame Mask AP. The filter reduces the clean-to-adversarial drop in
all prefix settings, and nearly eliminates it when the attacked prefix is long.
However, this reduction comes with a large clean-performance cost: for example,
at $1/14$, clean Mask AP decreases from 47.45 without filtering to 31.44 with
filtering. The filter therefore behaves as an aggressive post-processing rule
rather than a robust recovery mechanism.

\begin{table}[t]
\centering
\small
\setlength{\tabcolsep}{5pt}

\begin{tabular}{lcccccc}
\toprule
& \multicolumn{3}{c}{No filter} & \multicolumn{3}{c}{Temporal filter} \\
\cmidrule(lr){2-4}\cmidrule(lr){5-7}
Attack/eval & Clean & Adv. & Drop & Clean & Adv. & Drop \\
\midrule
$1/14$ & 47.45 & 43.45 & 4.00 & 31.44 & 28.90 & 2.54 \\
$7/8$ & 48.26 & 42.95 & 5.31 & 31.41 & 30.86 & 0.55 \\
$14/1$ & 53.94 & 43.58 & 10.36 & 38.32 & 37.21 & 1.11 \\
\bottomrule
\end{tabular}
\caption{Temporal consistency filtering for image-to-video transfer. Scores are tail-frame Mask AP percentages on SACo-VEval. Drop is clean minus adversarial.}
\label{tab:defense_video}
\end{table}

Overall, these defenses do not invalidate the main UCD finding. Test-time prompt
ensembling and temporal filtering can reduce the measured attack drop, but only
by lowering clean concept-segmentation quality. Head fine-tuning is less
destructive, yet UCD remains effective against both the original and fine-tuned
models. Robust defenses for SAM3 therefore need to preserve the presence-gated
concept decision rather than simply pruning uncertain predictions.

\section{Related Work}

\subsection{Segment Anything Models}

SAM introduced class-agnostic promptable segmentation from points, boxes, masks,
and automatic prompts \cite{kirillov2023segment}, and has often served as the
segmentation prior in detector- or grounding-driven open-vocabulary pipelines
\cite{2021Grounded,2024Grounded}. SAM2~\cite{ravi2025sam} keeps this geometric-prompt interface
while extending it to video with streaming memory. SAM3~\cite{carion2025sam}
instead moves to Promptable Concept Segmentation (PCS). For UCD,
the change is SAM3's presence-token gate, which makes concept existence a
shared post-processing decision and exposes the attack surface targeted in this
work.

\subsection{Adversarial Attacks on SAM Series}

Adversarial attacks~\cite{2013Intriguing,2014Explaining,2017Towards2,2017Towards,2015DeepFool} add small perturbations to induce incorrect predictions,
with universal adversarial perturbations learning one reusable perturbation
across inputs \citep{2017Universal}.
Several studies~\cite{zhou2024darksam,2024Transferable,2024Unsegment,2026Region,2024Segment,10.1109/TMM.2024.3521769,10516626,han2024sammeetsuapattacking} have analyzed or attacked SAM-series models. UAP-SAM2~\cite{zhou2026vanish} learns cross-prompt
universal perturbations under video memory with target scanning and dual
semantic deviation. Unlike these attacks, UCD trains on positive (image, NP) pairs and directly targets SAM3.

\section{Conclusion}

This paper studies the adversarial robustness of SAM3 image segmentation and
identifies the presence-gated concept decision as a new universal attack
surface. We proposed UCD, a cross-concept universal adversarial perturbation.
Experiments across five image benchmarks show that UCD consistently outperforms all attack baselines, and ablations verify
the contribution of each component. Our
defense-oriented evaluation further suggests that simple prompt ensembling,
head fine-tuning, and temporal filtering are insufficient to restore robust
concept segmentation without cost.

\bibliography{aaai2027}

@inproceedings{kirillov2023segment,
  title={Segment anything},
  author={Kirillov, Alexander and Mintun, Eric and Ravi, Nikhila and Mao, Hanzi and Rolland, Chloe and Gustafson, Laura and Xiao, Tete and Whitehead, Spencer and Berg, Alexander C and Lo, Wan-Yen and others},
  booktitle={Proceedings of the IEEE/CVF international conference on computer vision},
  pages={4015--4026},
  year={2023}
}

@inproceedings{ravi2025sam,
  title={Sam 2: Segment anything in images and videos},
  author={Ravi, Nikhila and Gabeur, Valentin and Hu, Yuan-Ting and Hu, Ronghang and Ryali, Chaitanya and Ma, Tengyu and Khedr, Haitham and R{\"a}dle, Roman and Rolland, Chloe and Gustafson, Laura and others},
  booktitle={International Conference on Learning Representations},
  volume={2025},
  pages={28085--28128},
  year={2025}
}

@article{carion2025sam,
  title={Sam 3: Segment anything with concepts},
  author={Carion, Nicolas and Gustafson, Laura and Hu, Yuan-Ting and Debnath, Shoubhik and Hu, Ronghang and Suris, Didac and Ryali, Chaitanya and Alwala, Kalyan Vasudev and Khedr, Haitham and Huang, Andrew and others},
  journal={arXiv preprint arXiv:2511.16719},
  year={2025}
}

@article{qiao2023robustness,
  title={Robustness of sam: Segment anything under corruptions and beyond},
  author={Qiao, Yu and Zhang, Chaoning and Kang, Taegoo and Kim, Donghun and Zhang, Chenshuang and Hong, Choong Seon},
  journal={arXiv preprint arXiv:2306.07713},
  year={2023}
}

@inproceedings{long2025robust,
  title={Robust SAM: on the adversarial robustness of vision foundation models},
  author={Long, Jiahuan and Xu, Zhengqin and Jiang, Tingsong and Yao, Wen and Jia, Shuai and Ma, Chao and Chen, Xiaoqian},
  booktitle={Proceedings of the AAAI Conference on Artificial Intelligence},
  volume={39},
  number={6},
  pages={5775--5783},
  year={2025}
}

@article{zhou2024darksam,
  title={Darksam: Fooling segment anything model to segment nothing},
  author={Zhou, Ziqi and Song, Yufei and Li, Minghui and Hu, Shengshan and Wang, Xianlong and Zhang, Leo Yu and Yao, Dezhong and Jin, Hai},
  journal={Advances in Neural Information Processing Systems},
  volume={37},
  pages={49859--49880},
  year={2024}
}

@article{zhang2023attack,
  title={Attack-sam: Towards evaluating adversarial robustness of segment anything model},
  author={Zhang, Chenshuang and Zhang, Chaoning and Kang, Taegoo and Kim, Donghun and Bae, Sung-Ho and Kweon, In So},
  journal={arXiv preprint arXiv:2305.00866},
  volume={1},
  number={3},
  pages={5},
  year={2023}
}

@article{zhou2026vanish,
  title={Vanish into thin air: Cross-prompt universal adversarial attacks for sam2},
  author={Zhou, Ziqi and Hu, Yifan and Song, Yufei and Li, Zijing and Hu, Shengshan and Zhang, Leo Yu and Yao, Dezhong and Zheng, Long and Jin, Hai},
  journal={Advances in Neural Information Processing Systems},
  volume={38},
  pages={142839--142861},
  year={2026}
}

@inproceedings{radford2021learning,
  title={Learning transferable visual models from natural language supervision},
  author={Radford, Alec and Kim, Jong Wook and Hallacy, Chris and Ramesh, Aditya and Goh, Gabriel and Agarwal, Sandhini and Sastry, Girish and Askell, Amanda and Mishkin, Pamela and Clark, Jack and others},
  booktitle={International conference on machine learning},
  pages={8748--8763},
  year={2021},
  organization={PmLR}
}

@article{bolya2026perception,
  title={Perception encoder: The best visual embeddings are not at the output of the network},
  author={Bolya, Daniel and Huang, Po-Yao and Sun, Peize and Cho, Jang Hyun and Madotto, Andrea and Wei, Chen and Ma, Tengyu and Zhi, Jiale and Rajasegaran, Jathushan and Bangalath, Hanoona and others},
  journal={Advances in Neural Information Processing Systems},
  volume={38},
  pages={60884--60937},
  year={2026}
}

@article{kingma2014adam,
  title={Adam: A method for stochastic optimization},
  author={Kingma, Diederik P and Ba, Jimmy},
  journal={arXiv preprint arXiv:1412.6980},
  year={2014}
}

@inproceedings{gupta2019lvis,
  title={Lvis: A dataset for large vocabulary instance segmentation},
  author={Gupta, Agrim and Dollar, Piotr and Girshick, Ross},
  booktitle={Proceedings of the IEEE/CVF conference on computer vision and pattern recognition},
  pages={5356--5364},
  year={2019}
}

@inproceedings{kazemzadeh2014referitgame,
  title={Referitgame: Referring to objects in photographs of natural scenes},
  author={Kazemzadeh, Sahar and Ordonez, Vicente and Matten, Mark and Berg, Tamara},
  booktitle={Proceedings of the 2014 conference on empirical methods in natural language processing (EMNLP)},
  pages={787--798},
  year={2014}
}

@inproceedings{wu2020phrasecut,
  title={Phrasecut: Language-based image segmentation in the wild},
  author={Wu, Chenyun and Lin, Zhe and Cohen, Scott and Bui, Trung and Maji, Subhransu},
  booktitle={Proceedings of the IEEE/CVF Conference on Computer Vision and Pattern Recognition},
  pages={10216--10225},
  year={2020}
}

@article{kuznetsova2020open,
  title={The open images dataset v4: Unified image classification, object detection, and visual relationship detection at scale},
  author={Kuznetsova, Alina and Rom, Hassan and Alldrin, Neil and Uijlings, Jasper and Krasin, Ivan and Pont-Tuset, Jordi and Kamali, Shahab and Popov, Stefan and Malloci, Matteo and Kolesnikov, Alexander and others},
  journal={International journal of computer vision},
  volume={128},
  number={7},
  pages={1956--1981},
  year={2020},
  publisher={Springer}
}

@inproceedings{shen2024practical,
  title={Practical region-level attack against segment anything models},
  author={Shen, Yifan and Li, Zhengyuan and Wang, Gang},
  booktitle={Proceedings of the IEEE/CVF Conference on Computer Vision and Pattern Recognition},
  pages={194--203},
  year={2024}
}

@article{2020Universal,
  title={Universal Adversarial Attack Via Enhanced Projected Gradient Descent},
  author={ Deng, Yingpeng  and  Karam, Lina J. },
  journal={IEEE},
  year={2020},
}

@article{2024Transferable,
  title={Transferable Adversarial Attacks on SAM and Its Downstream Models},
  author={ Xia, Song  and  Yang, Wenhan  and  Yu, Yi  and  Lin, Xun  and  Ding, Henghui  and  Duan, Ling Yu  and  Jiang, Xudong },
  year={2024},
}

@article{2017Universal,
  title={Universal adversarial perturbations},
  author={ Moosavi-Dezfooli, Seyed Mohsen  and  Fawzi, Alhussein  and  Fawzi, Omar  and  Frossard, Pascal },
  journal={IEEE},
  year={2017},
}

@article{2021Grounded,
  title={Grounded Language-Image Pre-training},
  author={ Li, Liunian Harold  and  Zhang, Pengchuan  and  Zhang, Haotian  and  Yang, Jianwei  and  Li, Chunyuan  and  Zhong, Yiwu  and  Wang, Lijuan  and  Yuan, Lu  and  Zhang, Lei  and  Hwang, Jenq Neng },
  year={2021},
}

@article{2024Grounded,
  title={Grounded SAM: Assembling Open-World Models for Diverse Visual Tasks},
  author={ Ren, Tianhe  and  Liu, Shilong  and  Zeng, Ailing  and  Lin, Jing  and  Li, Kunchang  and  Cao, He  and  Chen, Jiayu  and  Huang, Xinyu  and  Chen, Yukang  and  Yan, Feng },
  year={2024},
}

@article{2017Towards,
  title={Towards Evaluating the Robustness of Neural Networks},
  author={ Carlini, Nicholas  and  Wagner, David },
  journal={IEEE},
  year={2017},
}

@article{2015DeepFool,
  title={DeepFool: a simple and accurate method to fool deep neural networks},
  author={ Moosavi-Dezfooli, Seyed Mohsen  and  Fawzi, Alhussein  and  Frossard, Pascal },
  journal={IEEE},
  year={2015},
}

@article{2024Unsegment,
  title={Unsegment Anything by Simulating Deformation},
  author={ Lu, Jiahao  and  Yang, Xingyi  and  Wang, Xinchao },
  journal={IEEE},
  year={2024},
}

@article{2026Region,
  title={Region-guided attack on the segment anything model},
  author={ Liu, Xiaoliang  and  Shen, Furao  and  Zhao, Jian },
  journal={Neural Networks},
  volume={193},
  number={c},
  pages={108058},
  year={2026},
}

@article{2024Segment,
  title={Segment Shards: Cross-Prompt Adversarial Attacks against the Segment Anything Model},
  author={ Huang, Shize  and  Fan, Qianhui  and  Zhang, Zhaoxin  and  Liu, Xiaowen  and  Song, Guanqun  and  Qin, Jinzhe },
  journal={Applied Sciences (2076-3417)},
  volume={14},
  number={8},
  year={2024},
}

@article{2013Intriguing,
  title={Intriguing properties of neural networks},
  author={ Szegedy, Christian  and  Zaremba, Wojciech  and  Sutskever, Ilya  and  Bruna, Joan  and  Erhan, Dumitru  and  Goodfellow, Ian  and  Fergus, Rob },
  journal={Computer Science},
  year={2013},
}

@article{2014Explaining,
  title={Explaining and Harnessing Adversarial Examples},
  author={ Goodfellow, Ian J.  and  Shlens, Jonathon  and  Szegedy, Christian },
  journal={Computer Science},
  year={2014},
}

@article{2017Towards2,
  title={Towards Deep Learning Models Resistant to Adversarial Attacks},
  author={ Madry, Aleksander  and  Makelov, Aleksandar  and  Schmidt, Ludwig  and  Tsipras, Dimitris  and  Vladu, Adrian },
  year={2017},
}

@INPROCEEDINGS{10516626,
  author={Croce, Francesco and Hein, Matthias},
  booktitle={2024 IEEE Conference on Secure and Trustworthy Machine Learning (SaTML)}, 
  title={Segment (Almost) Nothing: Prompt-Agnostic Adversarial Attacks on Segmentation Models}, 
  year={2024},
  volume={},
  number={},
  pages={425-442},
  doi={10.1109/SaTML59370.2024.00028}}

@misc{han2024sammeetsuapattacking,
      title={SAM Meets UAP: Attacking Segment Anything Model With Universal Adversarial Perturbation}, 
      author={Dongshen Han and Chaoning Zhang and Sheng Zheng and Chang Lu and Yang Yang and Heng Tao Shen},
      year={2024},
      eprint={2310.12431},
      archivePrefix={arXiv},
      primaryClass={cs.CV},
      url={https://arxiv.org/abs/2310.12431}, 
}

@article{10.1109/TMM.2024.3521769,
author = {Zheng, Sheng and Zhang, Chaoning and Hao, Xinhong},
title = {Black-Box Targeted Adversarial Attack on Segment Anything (SAM)},
year = {2025},
issue_date = {2025},
publisher = {IEEE Press},
volume = {27},
issn = {1520-9210},
url = {https://doi.org/10.1109/TMM.2024.3521769},
doi = {10.1109/TMM.2024.3521769},
journal = {Trans. Multi.},
month = jan,
pages = {1901–1913},
numpages = {13}
}

\end{document}